\pdfoutput=1

\documentclass[11pt]{article}

\usepackage[preprint]{acl}

\usepackage{times}
\usepackage{latexsym}

\usepackage[T1]{fontenc}

\usepackage[utf8]{inputenc}

\usepackage{microtype}

\usepackage{inconsolata}

\usepackage{graphicx}

\usepackage{amsmath}
\usepackage{booktabs} 
\usepackage[most]{tcolorbox}
\usepackage{multirow}
\usepackage{enumitem}
\usepackage{threeparttable}
\usepackage{geometry}
\usepackage{array}
\usepackage{caption}
\usepackage{tabularx}
\usepackage{amssymb}

\usepackage{adjustbox}

\renewcommand{\arraystretch}{1.2} 

\newcommand{\clip}{\operatorname{clip}}
\newcommand{\rclip}{\clip(\rho_i,1-\varepsilon,1+\varepsilon)}
\newcommand{\KL}{\mathrm{KL}}

\usepackage[utf8]{inputenc}
\usepackage{tcolorbox}
\usepackage{lipsum} 
\usepackage{hyperref}
\usepackage{adjustbox}

\tcbset{
  rethinkstyle/.style={
    colback=gray!5!white,
    colframe=black!50,
    boxrule=0.4pt,
    arc=2pt,
    fonttitle=\bfseries,
    title=Rethink Prompt Template,
    fontupper=\footnotesize,
    left=6pt,
    right=6pt,
    top=6pt,
    bottom=6pt,
    sharp corners,
    enhanced,
    breakable,
    width=\textwidth
  }
}

\tcbset{
  reasonstyle/.style={
    colback=gray!5!white,
    colframe=black!50,
    boxrule=0.4pt,
    arc=2pt,
    fonttitle=\bfseries,
    title=DeepSeek-R1 node classification reasoning,
    fontupper=\footnotesize,
    left=6pt,
    right=6pt,
    top=6pt,
    bottom=6pt,
    sharp corners,
    enhanced,
    breakable,
  }
}

\title{\textsc{Graph-R1}: Incentivizing the Zero-Shot Graph Learning Capability in LLMs via Explicit Reasoning}

\author{
 \textbf{Yicong Wu\thanks{Equal contribution.}},
 \textbf{Guangyue Lu\footnotemark[1]},
 \textbf{Yuan Zuo\thanks{Corresponding author.}},
 \textbf{Huarong Zhang},
 \textbf{Junjie Wu},
\\
 MIIT Key Laboratory of Data Intelligence and Management, Beihang University
\\
 \small{
    \{2408312, lugybuaa, zuoyuan, 18377472, wujj\}@buaa.edu.cn}
 }

\begin{document}
\maketitle

\begin{abstract}
Generalizing to unseen graph tasks without task-specific supervision remains challenging. Graph Neural Networks (GNNs) are limited by fixed label spaces, while Large Language Models (LLMs) lack structural inductive biases. Recent advances in Large Reasoning Models (LRMs) provide a zero-shot alternative via explicit, long chain-of-thought reasoning. Inspired by this, we propose a GNN-free approach that reformulates graph tasks—node classification, link prediction, and graph classification—as textual reasoning problems solved by LRMs. We introduce the first datasets with detailed reasoning traces for these tasks and develop \textsc{Graph-R1}, a reinforcement learning framework that leverages task-specific rethink templates to guide reasoning over linearized graphs. Experiments demonstrate that \textsc{Graph-R1} outperforms state-of-the-art baselines in zero-shot settings, producing interpretable and effective predictions. Our work highlights the promise of explicit reasoning for graph learning and provides new resources for future research. Codes are available at \url{https://anonymous.4open.science/r/emnlp_submission-FDF0}.
\end{abstract}

\section{Introduction}

Zero-shot learning in graph machine learning aims to solve tasks in unseen label spaces or domains without any task-specific supervision.  
While graph neural networks (GNNs) perform well when ample labeled data are available, their generalization ability sharply deteriorates under distribution shifts or in new label spaces—unless expensive fine-tuning is applied~\citep{juMultitaskSelfsupervisedGraph2023}.  
Prompt-based GNN variants~\citep{liuGraphPromptUnifyingPretraining2023a, sunAllOneMultiTask2023a}, inspired by advances in natural language processing (NLP), offer partial mitigation; however, their fixed, task-specific output heads still hinder true zero-shot generalization.

Large language models (LLMs) offer a complementary and promising alternative.  
A straightforward approach flattens the graph into a textual sequence and feeds it to an LLM~\citep{chenExploringPotentialLarge2024a, guoGPT4GraphCanLarge2023b, wangCanLanguageModels2023a, liuEvaluatingLargeLanguage2023a}. However, this often yields suboptimal results due to the lack of structural inductive bias essential for effective graph reasoning~\citep{huangCanLLMsEffectively2024}.  
Recent efforts have sought to more tightly integrate GNNs with LLMs.  
One line of work retains the GNN as the predictor while using the LLM to generate auxiliary signals, such as synthetic labels or node descriptions~\citep{yeLanguageAllGraph2024, yuLeveragingLargeLanguage2025, xiaOpenGraphOpenGraph2024, chenLabelfreeNodeClassification2024}. Yet, these methods still rely on rigid GNN heads and require retraining for each task.  
Another approach delegates prediction to the LLM while incorporating structural signals from a frozen GNN via cross-modal projection~\citep{tangGraphGPTGraphInstruction2024a,heUniGraphLearningUnified2025,wangLLMsZeroshotGraph2024b}. Unfortunately, the separation of training between components results in weak task conditioning and limited transferability.  
More tightly coupled methods—such as GOFA~\citep{kongGOFAGenerativeOneForAll2024a}—inject GNN features directly into the LLM token stream at inference time. While this improves zero-shot accuracy, it introduces substantial computational overhead and still struggles with generalization across tasks and domains.

\paragraph{From \emph{graph structure} to text-based \emph{reason-then-predict}.}
Recent advances in Large Reasoning Models (LRMs) (e.g., \textsc{DeepSeek-R1}~\citep{deepseek-aiDeepSeekR1IncentivizingReasoning2025}) renew our interest in the graph-to-text paradigm, driven by their ability to generate explicit reasoning processes.
These models can potentially compensate for the lack of handcrafted structural priors and offer an interpretable, zero-shot-capable alternative for graph learning.  
Crucially, many canonical graph tasks—such as link prediction, edge classification and node or graph-level classification—can be naturally reformulated as short sequences of relational reasoning steps, once the graph is linearized into text.  
Prompting a reasoning-capable LLM to generate such chains of deduction effectively replaces the opaque feedforward process of a GNN with a transparent \emph{reason-then-predict} pipeline. 
This shift offers two key advantages: improved generalization under distribution shift (since the model must justify each step rather than memorize patterns), and human-interpretable rationales for every prediction.  
Progress in this direction, however, is currently bottle-necked by the lack of (i) a benchmark that evaluates both reasoning and prediction across a wide range of graph tasks, and (ii) a model that fully commits to this pure, GNN-free paradigm.

\paragraph{Our solution.}
To tackle the above challenges, we present the first dataset that simultaneously spans node classification, link prediction, and graph classification—each annotated with explicit chains of thought.  
Leveraging this resource, we develop \textsc{Graph-R1}—a reinforcement-learning-based, purely graph-to-text framework that promotes LLM reasoning for cross-task, cross-domain zero-shot prediction without any GNN component.  
We further design a \emph{rethink reasoning template} specialized for graph prediction.  
Comprehensive experiments show that \textsc{Graph-R1} matches or exceeds strong baselines, particularly in challenging transfer scenarios, highlighting the value of explicit reasoning for graph learning. Our contributions are threefold:
\begin{itemize}
\item We construct the first reasoning dataset tailored for graph machine learning tasks, with detailed reasoning traces.
\item We propose \textsc{Graph-R1}, a reasoning-augmented LLM that improves generalization and transferability via reinforcement learning with task-specific rethink templates.
\item Extensive experiments show state-of-the-art zero-shot performance across diverse graph tasks, demonstrating the impact of explicit reasoning in LLM-based graph prediction.
\end{itemize}

\section{Methods}
We present \textsc{Graph-R1}, a \emph{graph-to-text} framework for cross-domain and cross-task graph reasoning.  
The method is built on three principal modules:  
(i) reasoning-centric graph dataset construction,
(ii) reasoning knowledge acquisition via full-parameter fine-tuning, and
(iii) reinforcement learning–based reasoning enhancement.

\subsection{Notations}

Let $\tau\!\in\!\mathcal{T}$ denote a graph-learning task type (e.g., node classification, link prediction).  
A graph is written as  
$
G=(\mathcal{V},\mathcal{E},\mathbf{A},\mathbf{X}),
$  
where $\mathcal{V}=\{v_1,\dots,v_{|\mathcal{V}|}\}$ and $\mathcal{E}=\{e_1,\dots,e_{|\mathcal{E}|}\}$ are the node and edge sets, respectively.  
The adjacency matrix $\mathbf{A}^G\!\in\!\{0,1\}^{|\mathcal{V}|\times|\mathcal{V}|}$ satisfies $A_{ij}^G=1$ iff $(v_i,v_j)\!\in\!\mathcal{E}$.  
Textual attributes are stored in $\mathbf{X}$: $\mathbf{x}(v_i)$ denotes the textual description of node $v_i$, and $\mathbf{x}(e_{ij})$ denotes that of edge $e_{ij}$, which connects nodes $v_i$ and $v_j$.
For a graph–task pair $(G,\tau)$ we design a prompt template $P_{G,\tau}$ that linearises $G$ and specifies the task requirements.

\begin{figure}[t]
\centering
\begin{tcolorbox}[reasonstyle]

\textcolor{black!70}{\textbf{Reasoning Traces:}} 

\vspace{0.5em}
Okay, Let's try to figure out how to classify \texttt{node3}.
So, the task is to determine the correct category for the essay represented by\ldots 

\vspace{0.5em}
\textcolor{cyan!70!black}{First}, I need to look at \texttt{node3}'s own description. It says: ``Practical lattice-coded PNC\ldots 

\vspace{0.5em}
\textcolor{cyan!70!black}{Second}, the connections suggest that \texttt{node3} is part of a network of papers related to physical-layer network coding\ldots 

\vspace{0.5em}
\textcolor{cyan!70!black}{Wait}, but the connections listed are \texttt{node3}-\texttt{node4}, \texttt{node3}-\texttt{node5}\ldots 

\vspace{0.5em}
\textcolor{cyan!70!black}{Therefore}, based on the above reasoning, the primary category for \texttt{node3} is likely \textbf{cs.NI}. 

\vspace{1em}
\texttt{Answer:} \textbf{cs.NI} 

\texttt{Brief\_reasoning:} 
\textcolor{gray!80!black}{\texttt{node3} is best categorized under \texttt{cs.NI} due to its focus on practical lattice-coded physical-layer network coding, with no meaningful structural or semantic connections to other domains.}

\end{tcolorbox}
\caption{An illustration of explicit reasoning traces produced by \textsc{DeepSeek-R1} for node classification.}
\label{fig:reasoning}
\vspace{-0.3cm}
\end{figure}

\subsection{Graph-Reasoning Data Curation}\label{sec:datamethod}

To investigate \emph{reason-then-predict} graph learning, we construct the first dataset featuring explicit, detailed reasoning traces across multiple graph tasks. 

\paragraph{Dataset and task selection.}
We sample 11 representative datasets from five domains—\textit{citation networks, e-commerce, social media, molecular graphs}, and \textit{knowledge graphs}.  
Together they cover node, edge, and graph-level tasks (node classification, link prediction, graph classification, edge classification), ensuring broad coverage for evaluating graph reasoning.

\paragraph{Graph-to-text augmentation.}
Unlike prior work that tokenizes structural features using GNN encoders, we revisit the pure graph-to-text paradigm. Taking node-level tasks as an example, for a target node $v_i$, we extract its $h$-hop subgraph and describe all node features $T_i=\{\mathbf{x}(v_j)\mid j\in\mathcal{N}(i)\cup\{i\}\}$, and edge relations $E_i=\{\mathbf{x}(e_{jk})\mid v_j,v_k\in\mathcal{N}(i)\cup\{i\}\}$ within the subgraph using natural language, where $\mathcal{N}(i)$ is the neighborhood of $v_i$.
To maintain input tractability for large graphs with verbose node texts (e.g., citation networks with titles and abstracts), we apply \textsc{DeepSeek-V3} for automatic summarization. Prompt templates are provided in Appendix~\ref{appendix:prompt}.

\paragraph{Reasoning-trace extraction.}  
A distinctive feature of our dataset construction is the inclusion of explicit reasoning traces for each answer. Specifically, each subgraph query $Q_i$ consists of node features $T_i$, edge relations $E_i$, and a prompt template $P_{G, \tau}$ tailored to the graph structure $G$ and task type $\tau$, serving as input to the LLM. We then input $Q_i$ into \textsc{DeepSeek-R1} to generate an explicit reasoning trace $R_i$ and a final prediction $Y_i$, as illustrated in  Figure~\ref{fig:reasoning}. Formally, this process can be represented as:
\[
Q_i \rightarrow (Y_i, R_i).
\]

\paragraph{Quality control.}
We apply a three-stage filtering process:
\begin{enumerate}[itemsep=2pt, parsep=0pt, topsep=2pt]
    \item \textbf{Information sufficiency}: remove isolated nodes and trivial subgraphs.  
    \item \textbf{Answer validity}: discard samples where the predicted answer $Y_i$ mismatches the gold label or contains sensitive content.  
    \item \textbf{Rationale coherence}: retain only rationales that exhibit reasonable length and logical consistency.  
\end{enumerate}
The final corpus contains 10,000 graph reasoning examples across multiple domains and tasks, each paired with an explicit chain-of-thought explanation.

\begin{figure}[t]
  \centering
  \includegraphics[width=\columnwidth]{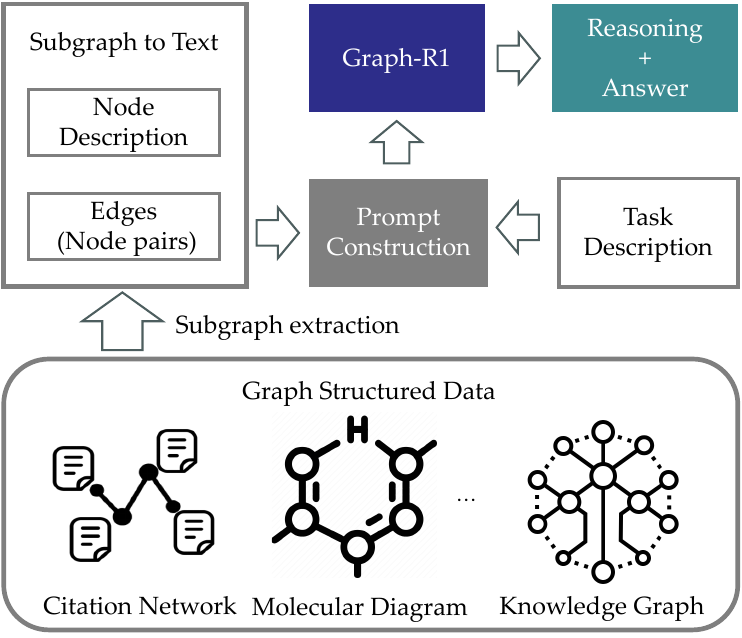}
  \caption{\textsc{Graph-R1} framework. Graphs are linearized into a graph description language, and a task-aware prompt guides the LLM to produce explicit reasoning and the final answer.}
  \label{fig:init vs sft}
\end{figure}

\subsection{Graph-R1}

Building on the graph–reasoning corpus described above, we develop \textsc{Graph-R1}, an LLM-based framework for solving graph machine learning tasks through explicit reasoning.  
Training proceeds in two stages: (1) joint instruction tuning across multiple tasks and domains, and (2) reinforcement learning to refine reasoning quality.  
To support smaller LLM backbones, we introduce a \emph{rethink} template that encourages deeper semantic and structural analysis, leading to more robust and interpretable multi-step deductions.  
This pipeline enables \textsc{Graph-R1} to advance zero-shot graph reasoning with large language models.

\begin{figure*}[t!]
\centering
\begin{tcolorbox}[rethinkstyle, width=\linewidth, sharp corners, boxrule=0.5pt]
\textcolor{black!70}{Question:} (will be dynamically filled) \\
You must conduct reasoning inside \textcolor{blue!70!black}{\texttt{<think>}}...\textcolor{blue!70!black}{\texttt{</think>}}. Inside it, you should include: \\
- Include topological analysis in \textcolor{violet!80!black}{\texttt{<structure>}}...\textcolor{violet!80!black}{\texttt{</structure>}} \\
- Include semantic interpretation in \textcolor{cyan!70!black}{\texttt{<semantic>}}...\textcolor{cyan!70!black}{\texttt{</semantic>}} \\
- Provide three candidate answers in \textcolor{orange!60!black}{\texttt{<comprehensive>}}...\textcolor{orange!60!black}{\texttt{</comprehensive>}} \\
- Re-evaluate each candidate in \textcolor{magenta!50!black}{\texttt{<rethink>}}...\textcolor{magenta!50!black}{\texttt{</rethink>}} \\

Your response must follow this format: \\
\textcolor{blue!70!black}{\texttt{<think>}} \\
\quad \textcolor{violet!80!black}{\texttt{<structure>}}Structure analysis here\textcolor{violet!80!black}{\texttt{</structure>}} \\
\quad \textcolor{cyan!70!black}{\texttt{<semantic>}}Semantic analysis here\textcolor{cyan!70!black}{\texttt{</semantic>}} \\
\quad \textcolor{orange!60!black}{\texttt{<comprehensive>}}List candidate answers and brief reasoning\textcolor{orange!60!black}{\texttt{</comprehensive>}} \\
\quad \textcolor{magenta!50!black}{\texttt{<rethink>}}Re-evaluate each candidate in depth\textcolor{magenta!50!black}{\texttt{</rethink>}} \\
Final reasoning and answer \\
\textcolor{blue!70!black}{\texttt{</think>}}\\
\texttt{Answer:} your\_answer \\
\texttt{Brief\_reasoning:} your\_brief\_reasoning
\end{tcolorbox}
\caption{Rethink Prompt Template. This structure-aware reasoning format is used during both training and inference.}
\label{fig:rethink-template}
\vspace{-0.2cm}
\end{figure*}

\subsubsection{Reasoning Knowledge Learning via Full-Parameter Fine-Tuning}

In Phase~1, we perform joint instruction tuning across node-, edge-, and graph-level tasks from diverse domains, transferring the general reasoning capabilities of \textsc{DeepSeek-R1} to the graph setting and leveraging multi-task synergies.

We adopt full-parameter supervised fine-tuning using the standard language modeling loss.  
Given a graph query $Q_i$—comprising textual node and edge features $\{T_i, E_i\}$ and a prompt $P_{G,\tau}$—the model is trained to generate both the reasoning trace $R_i$ and the final answer $Y_i$:
\begin{equation}
\mathcal{L}(\theta) = -\frac{1}{N} \sum_{i=1}^{N} \log p_{\theta}\bigl(R_i, Y_i \mid Q_i\bigr),
\end{equation}
where $N$ is the number of training examples and $\theta$ denotes the model parameters.  
The model thus learns to map graph-structured prompts to coherent reasoning traces and accurate solutions.  
Exposure to a broad range of tasks enhances generalization and promotes transferable reasoning abilities.  
Detailed training configurations are provided in Appendix~\ref{appendix:experiment}.

\subsubsection{Reinforcement-Learning-Based Reasoning Enhancement}

In Phase~2, we refine the instruction-tuned model using Group Relative Policy Optimization (GRPO)~\citep{shaoDeepSeekMathPushingLimits2024}, a reinforcement learning method that jointly incentivizes answer correctness and the logical coherence of the reasoning trace—thereby enhancing generalization across graph tasks.

GRPO fine-tunes the supervised model using a reward signal that balances reasoning quality and prediction accuracy. Its training objective is:

\begin{subequations}\footnotesize
\begin{equation}
\label{eq:grpo_obj}
\begin{aligned}
\mathcal{J}(\theta)=
\mathbb{E}_{q\sim P(Q)}\frac{1}{g}\sum_{i=1}^{g}
\Bigl[&\min(\rho_i A_i,\rclip A_i)\\
      &-\beta\,\KL(\pi_\theta\|\pi_{\text{ref}})\Bigr].
\end{aligned}
\end{equation}
\begin{equation}
\label{eq:rho}
\rho_i=\frac{\pi_\theta(o_i\mid q)}{\pi_{\theta_{\text{old}}}(o_i\mid q)}.
\end{equation}
\begin{equation}
\label{eq:kl}
\KL(\pi_\theta\|\pi_{\text{ref}})=
\frac{\pi_{\text{ref}}(o_i\mid q)}{\pi_\theta(o_i\mid q)}-
\log\frac{\pi_{\text{ref}}(o_i\mid q)}{\pi_\theta(o_i\mid q)}-1.
\end{equation}
\end{subequations}
Here, $\pi_\theta$ and $\pi_{\theta_{\text{old}}}$ denote the current and previous policies, respectively; $q$ and $o_i$ represent the sampled question and its $i$-th response; and $g$ is the group size.  
The hyperparameters $\epsilon$ and $\beta$ control the clipping threshold and KL divergence penalty, respectively.  
The group-wise advantage is computed as:
\[
A_i = \frac{r_i - \operatorname{mean}(\{r_j\}_{j=1}^{g})}
           {\operatorname{std}(\{r_j\}_{j=1}^{g})},
\]
where $\{r_j\}$ are the rewards for the $g$ responses to the same question.

\paragraph{Rethink Template.}
Conventional prompting typically restricts reasoning to a single \texttt{<think>} block—effective for mathematical problems, but suboptimal for graph tasks where both structural and semantic information are critical, and labels can be ambiguous.  
To address this, we propose a \emph{rethink} reasoning template specifically designed for graph prediction tasks (see Figure~\ref{fig:rethink-template}).

Our revised template introduces a structured, multi-phase reasoning process:
\begin{itemize}[itemsep=2pt, parsep=0pt, topsep=2pt]
    \item \texttt{<structure>}: encourages explicit topological analysis;
    \item \texttt{<semantic>}: focuses on the interpretation of node/edge attributes;
    \item \texttt{<comprehensive>}: elicits multiple candidate answers to expose alternative hypotheses;
    \item \texttt{<rethink>}: revisits each candidate to encourage comparative and bidirectional evaluation.
\end{itemize}

This structure-aware prompting scheme enables tighter integration of topology and semantics, significantly improving RL performance on node classification, link prediction tasks and etc.

\paragraph{Reward Modeling.}
The reward function serves as the core training signal in reinforcement learning.  
Under the \emph{standard} \texttt{<think>} template, we adopt a simple reward scheme:
\begin{equation}
\footnotesize
\label{eq:normal_reward}
R=
\begin{cases}
1      & \text{if the answer is correct},\\
0.01   & \text{if the output is merely well-formatted},\\
0      & \text{otherwise}.
\end{cases}
\end{equation}

While sufficient for toy mathematical tasks, this coarse-grained feedback overlooks the rich intermediate reasoning required for graph-based problems.  
To address this, we design a more fine-grained reward for the \emph{rethink} template, which evaluates both the reasoning trace and the final answer.  
During the initial reasoning phase, the model lists multiple candidate answers; partial credit is assigned if the gold label appears among them:
\begin{equation}
\footnotesize
\label{eq:rethink_reward}
R=
\begin{cases}
1      & \text{if the final answer is correct},\\
0.3    & \text{if the correct answer appears in \texttt{<rethink>}},\\
0.01   & \text{if only the format is correct},\\
0      & \text{otherwise}.
\end{cases}
\end{equation}

Coupled with GRPO, this refined reward enables the model to learn richer and more reliable reasoning paths, leading to state-of-the-art zero-shot performance across all evaluated graph domains and tasks.

\begin{table*}[t]
\centering
\resizebox{\textwidth}{!}{%
\begin{tabular}{c|cccccccccc}
\toprule
 \textbf{Task} & \multicolumn{2}{c}{\textbf{Cora–Node}} & \multicolumn{2}{c}{\textbf{WikiCS}} 
 & \multicolumn{3}{c}{\textbf{Products}} & \textbf{Expla-Graph} & \textbf{Cora–Link} & \textbf{FB15K237} \\
\cmidrule(r){2-3} \cmidrule(r){4-5} \cmidrule(r){6-8} \cmidrule(r){9-9} \cmidrule(r){10-10} \cmidrule(r){11-11}
Way / Type & 7 & 2 & 10 & 5 & 47 & 10 & 5 & 2 & 2 & 10 \\ \midrule
Llama2-7B      & 47.92 & 73.45 & 40.10 & 58.77 & 27.65 & 58.71 & 64.33 & 57.76 & 48.15 & 48.32 \\
Mistral-7B     & 60.54 & 88.39 & 63.63 & 71.90 & 43.99 & 70.16 & 74.94 & 68.77 & 49.43 & 62.48 \\ \midrule
OFA            & 28.65 & 56.92 & 21.20 & 35.15 & 19.37 & 30.43 & 39.31 & 51.36 & 52.22 & --   \\
GraphGPT       & 44.65 & --    & --    & --    & 18.84 & --    & --    & --    & 50.74 & --   \\
UniGraph       & 69.53 & \textbf{89.74} & 43.45 & 60.23 & 38.45 & 66.07 & 75.73 & --    & --    & --   \\
ZeroG          & 64.21 & 87.83 & 31.26 & 48.25 & 31.24 & 51.24 & 71.29 & --    & --    & --   \\
LLaGA          & 51.85 & 62.73 & --    & --    & 23.10 & 34.15 & 39.72 & --    & \textbf{88.09} & --   \\
GOFA-T         & \underline{70.81} & 85.73 & 71.17 & \underline{80.93} & 54.60 & 79.33 & 87.13 & \underline{79.49} & 85.10 & 73.59 \\ 
GOFA-F         & 69.41 & 87.52 & \underline{68.84} & 80.52 & \underline{56.13} & \underline{80.03} & \underline{88.34} & 71.34 & \underline{86.31} & \textbf{80.69} \\ \midrule
\textbf{Graph-R1} & \textbf{71.53} & \underline{89.08} & \textbf{78.68} & \textbf{86.89} & \textbf{66.59} & \textbf{85.72} & \textbf{91.78} & \textbf{89.71} & \underline{86.31} & \underline{75.17} \\
\bottomrule
\end{tabular}}
\caption{Zero-shot accuracy (\%) across datasets. Best in \textbf{bold}; second best \underline{underlined}.}
\label{tab:main_result}
\end{table*}

\section{Experiments}
We begin by introducing the datasets used to train and evaluate \textsc{Graph-R1} (§\ref{sec:datasets}), followed by the baselines and experimental setup (§\ref{sec:setup}).  
We then present a comprehensive suite of experiments to assess the effectiveness and generalization of our method, focusing on the following questions:
\textbf{RQ1:} Does \textsc{Graph-R1} enable critical applications of general graph models, such as zero-shot learning?
\textbf{RQ2:} Can it generalize to unseen tasks and domains, including cross-task transfer?
\textbf{RQ3:} How do instruction tuning and the \emph{rethink} template contribute to generalization?
\textbf{RQ4:} How does \textsc{Graph-R1} compare to large reasoning models on graph tasks?

\subsection{Datasets}
\label{sec:datasets}
We evaluate \textsc{Graph-R1} on five benchmark datasets:
\begin{itemize}[leftmargin=1.5em, itemsep=2pt, parsep=0pt, topsep=2pt]
 \item Cora — citation network with node and link prediction tasks~\citep{wenAugmentingLowresourceText2023}.
  \item Products — e-commerce graph for node classification~\cite{heHarnessingExplanationsLLMtoLM2024}.
  \item WikiCS — Wikipedia graph with node classification~\citep{mernyeiWikiCSWikipediabasedBenchmark2020}.
  \item FB15K237 — knowledge graph for link prediction~\cite{liuOneAllTraining2024a}.
  \item Expla-Graph — synthetic graph reasoning benchmark~\citep{heGretrieverRetrievalaugmentedGeneration2024}.
\end{itemize}
All tasks are aligned with the evaluation protocol of GOFA~\citep{kongGOFAGenerativeOneForAll2024a}.  
To test cross-domain and cross-task generalization, we additionally evaluate on three unseen graph regression datasets—ESOL~\cite{withnallMatchedMolecularPair2018}, Lipo~\citep{wuMoleculeNetBenchmarkMolecular2017}, and Freesolv~\cite{casasnovasTheoreticalPKaCalculations2014}—which are not seen during either fine-tuning or reinforcement learning.  
This ensures a strict zero-shot cross-task setting.  
Dataset statistics and task details are provided in Appendix~\ref{appendix:dataset}.

\subsection{Experimental Setup}
\label{sec:setup}

\paragraph{Baselines.}
We compare against two groups of baselines:
\begin{itemize}[leftmargin=1.5em, itemsep=2pt, parsep=0pt, topsep=2pt]
    \item \textit{General-purpose LLMs}: LLaMA~2-7B~\citep{touvronLlama2Open2023}, Mistral-7B~\citep{jiangMistral7B2023}, and DeepSeek-R1-distilled-Qwen2.5-14B~\citep{deepseek-aiDeepSeekR1IncentivizingReasoning2025};
    \item \textit{Graph models leveraging LLMs}: OFA~\citep{liuOneAllTraining2024a}, GraphGPT~\citep{tangGraphGPTGraphInstruction2024a}, UniGraph~\citep{heUniGraphLearningUnified2025}, ZeroG~\citep{liZeroGInvestigatingCrossdataset2024a}, LLaGA~\citep{chenLLaGALargeLanguage2024b}, and GOFA~\citep{kongGOFAGenerativeOneForAll2024a}.
\end{itemize}
These baselines represent the current state-of-the-art in both general LLM and graph-specific LLM paradigms, providing a rigorous comparison for our proposed approach.

\paragraph{Implementation.}
We instantiate \textsc{Graph-R1} with DeepSeek-R1-distilled-Qwen2.5 models (14B).  
The model is first instruction-tuned and then further optimized with GRPO-based reinforcement learning on our graph reasoning dataset.  
All methods, including baselines, are evaluated under consistent zero-shot conditions and identical hardware.  
Hyperparameters are tuned based on validation performance.
Full training details, data splits, and evaluation metrics are available in Appendix~\ref{appendix:experiment}.

\subsection{Cross-Dataset Zero-Shot Generalization (RQ1)}
\label{sec:main_result}

To address RQ1, we run strict zero-shot evaluations on the GOFA-aligned benchmarks listed in §\ref{sec:datasets}.  
Table~\ref{tab:main_result} yields the following observations.
Generic LLMs such as Llama2-7B and Mistral-7B rely mainly on textual cues. They are competitive on node-classification datasets, where semantics dominate, but drop sharply on link-prediction tasks that require relational reasoning.
LLM-as-predictor models (GOFA, UniGraph) consistently surpass GNN-based hybrids (OFA, ZeroG). Encoding graph structure into the LLM token stream or feature space markedly improves cross-domain robustness.

\textsc{Graph-R1} attains the best accuracy on eight of ten settings and the second best on the rest, without any GNN encoder. Its graph-to-text reformulation plus reinforcement-learned reasoning allows the model to fuse topology and semantics purely in natural-language form, setting a new state of the art for zero-shot graph prediction.

\subsection{Cross-Task Zero-Shot Generalization (RQ2)}
\label{sec:cross_task}

To evaluate the model’s generalization ability in the zero-shot cross-task setting, we conduct a test in which the model is trained solely on classification-style tasks—node, edge, graph classification or link prediction—and is evaluated on \emph{unseen} graph regression tasks.
We compare \textsc{Graph-R1} with two representative LLM-based baselines, LLaGA and GOFA, both evaluated under the same zero-shot setting without access to regression training data.  
Results are shown in Table~\ref{tab:regression}.

\begin{table}[t]
\centering
\resizebox{0.35\textwidth}{!}{%
\begin{tabular}{lccc}
\toprule
\multirow{2}{*}{\textbf{Model}} & \multicolumn{3}{c}{\textbf{MAE} $\downarrow$} \\ \cmidrule(l){2-4}
& \textbf{ESOL} & \textbf{Lipo} & \textbf{FreeSolv} \\ \midrule
LLaGA        & 7.39 & 15.55 & 51.72 \\
GOFA         & \underline{4.93} & \textbf{1.36} & \underline{14.98} \\
\textsc{Graph-R1} & \textbf{1.72} & \underline{1.55} & \textbf{11.59} \\ \bottomrule
\end{tabular}}
\caption{Zero-shot graph regression results (lower MAE is better). Best in \textbf{bold}; second best \underline{underlined}.}
\label{tab:regression}
\end{table}

\textsc{Graph-R1} achieves the best performance on ESOL and FreeSolv and ranks second on Lipo, outperforming all baselines without any task-specific tuning.  
These results highlight its strong cross-task generalization—crucial for real-world deployment where labeled data are often scarce or unavailable.

\subsection{Ablation Study (RQ3)}
\label{sec:ablation}

To address \textbf{RQ3}, we ablate two key components of \textsc{Graph-R1}: instruction tuning and reinforcement learning (RL) with the \emph{rethink} template.  
Specifically, we compare four variants: (i) \emph{init} (the initial model without task-specific training), (ii) \emph{w/o RL} (instruction-tuned without RL), (iii) \emph{normal} (RL with the standard template), and (iv) the full \emph{Graph-R1} (RL with the rethink template).  

\begin{figure}[t!]
  \includegraphics[width=\columnwidth]{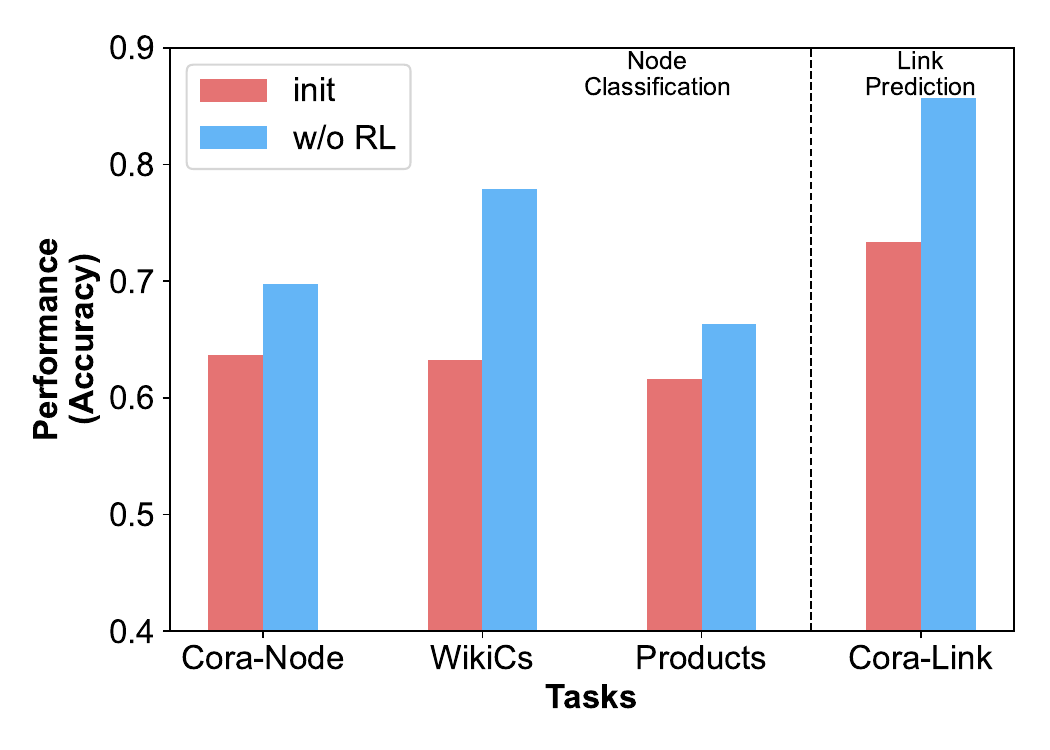}
  \caption{Comparison the results of initial model and instruction-tuned model.}
  \label{fig:init_vs_sft}
\end{figure}

\paragraph{Effect of Instruction Tuning.}  
As shown in Figure~\ref{fig:init_vs_sft}, instruction tuning alone consistently outperforms the initial model across all datasets.  
This demonstrates effective knowledge transfer and the distillation of graph reasoning capabilities from DeepSeek-R1 to our model, significantly enhancing its graph-specific inference performance.

\begin{figure}[t!]
  \includegraphics[width=\columnwidth]{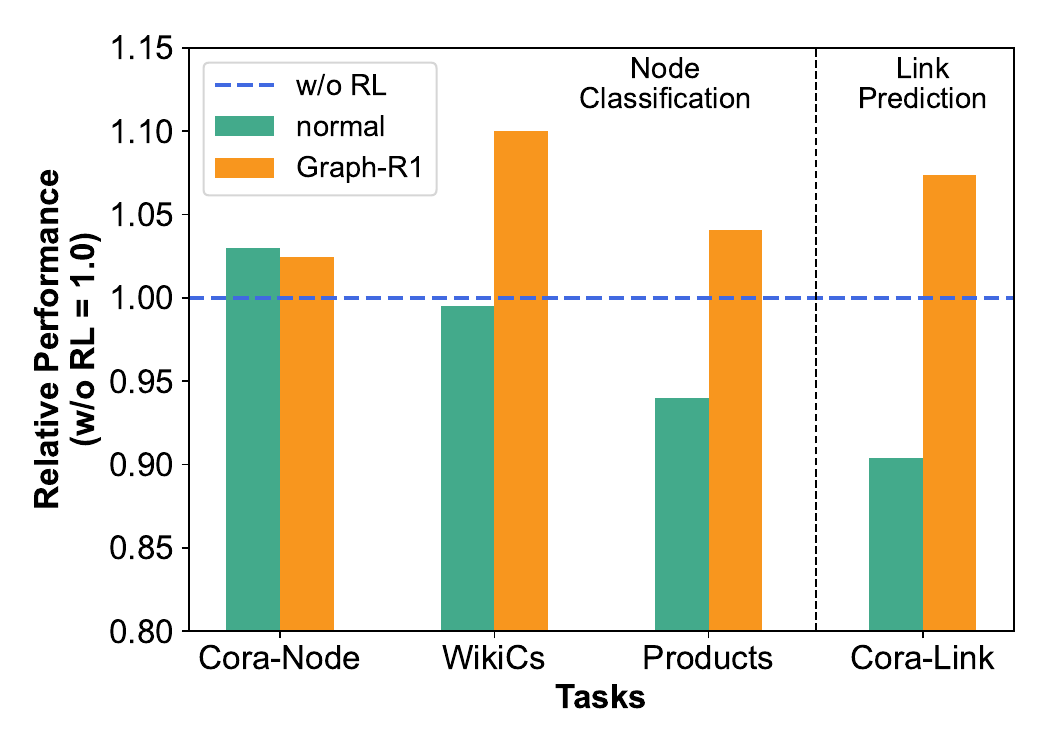}
  \caption{Comparison of RL performance with the standard template ("normal") and the rethink template ("rethink"), using the w/o RL result as the baseline (indicated by the blue dashed line at 1.0). All values are normalized with respect to the w/o RL baseline for each dataset.}
  \label{fig:norm_vs_rethink}
\end{figure}

\paragraph{Effect of RL with the Rethink Template.}  
To better illustrate the effect of reinforcement learning, we compare the \emph{normal} and \emph{Graph-R1} variants using the \emph{w/o RL} variant as the baseline. Results are presented in Figure~\ref{fig:norm_vs_rethink}, where the $y$-axis denotes normalized performance (i.e., the ratio of the performance to the w/o RL baseline).  
Applying RL with the standard template improves performance primarily on text- and logic-oriented tasks (e.g., Cora–Node, Expla-Graph), but leads to degradation on structure-heavy tasks such as Cora–Link and FB15K237, suggesting limited gains in structural reasoning.  
In contrast, RL with the rethink template yields consistent improvements across all tasks, underscoring its importance in enhancing both semantic and structural understanding, and thereby significantly boosting generalization.

\begin{table*}[t!]
    \centering
    \resizebox{\textwidth}{!}{ 
        \begin{tabular}{c|cccccc}
            \toprule
            \textbf{Task} & \textbf{Cora-Node} & \textbf{WikiCS} & \textbf{Products} & \textbf{ExplaGraphs} & \textbf{Cora-Link} & \textbf{FB15K237}\\ 
            
            \cmidrule(r){2-2} \cmidrule(r){3-3} \cmidrule(r){4-4} \cmidrule(r){5-5} \cmidrule(r){6-6} \cmidrule(r){7-7}
            
            Way/Type   & 7     & 10    & 47    & 2     & 2     & 10\\
            \midrule
            DeepSeek-R1-Distill-Qwen-14B & 60.67 & 69.33 & 57.33 & 81.33 & \underline{72.00} & 34.00\\
            DeepSeek-R1-671B  & \underline{68.67} & \underline{76.00} & \textbf{69.33} & \textbf{92.00} & 68.00 & \textbf{84.67}\\
            Graph-R1   & \textbf{72.67}  & \textbf{78.67} & \underline{65.33} & \underline{88.67} & \textbf{86.67} & \underline{72.00} \\ 
            \bottomrule
        \end{tabular}
    }
    \caption{Comparison between Graph-R1 and Large Reasoning Models (LRMs). Best in \textbf{bold}; second best \underline{underlined}.}
    \label{base model results}
\end{table*}

\subsection{Comparison with Large Reasoning Models (RQ4)}
\label{sec:vs_base}
Answering \textbf{RQ4} is crucial for understanding both the necessity of our two-phase training strategy and the effectiveness of \textsc{Graph-R1} as a general-purpose graph model.  
We compare \textsc{Graph-R1} with Large Reasoning Models (LRMs), including DeepSeek-R1 (671B) and its 14B distilled variant. All models are evaluated using identical input formats to ensure a fair comparison.  
Due to the high computational cost of DeepSeek-R1, we randomly sample 1,000 examples from the evaluation dataset introduced in Section~\ref{sec:datasets} for testing. The results on these samples are presented in Table~\ref{base model results}.

\textsc{Graph-R1} achieves substantial improvements over the 14B distilled model across all evaluated tasks.  
Moreover, in several cases, it matches or even surpasses the performance of DeepSeek-R1 (671B). 
These results provide strong evidence that our two-phase training strategy significantly enhances reasoning capabilities on graph tasks.

\subsection{Case Studies of \textsc{Graph-R1} Reasoning}
\label{sec:case}

To demonstrate the interpretability and reasoning capabilities of \textsc{Graph-R1}, we present two illustrative examples from its inference process on distinct graph tasks: node classification and link prediction.
Due to space constraints, the full case details are provided in Table~\ref{tab:case_studies} in Appendix~\ref{appendix:case}.
These examples highlight key aspects of the model’s reasoning process, including its ability to integrate structural and semantic information, comprehensively evaluate candidate options, and effectively verify hypotheses.

In node classification, the model showcased its ability to comprehensively evaluate multiple candidate categories by combining structural and semantic analyses, prioritizing the most relevant category, and systematically re-evaluating each candidate to confirm its conclusion. For link prediction, \textsc{Graph-R1} excelled in hypothesis testing during the rethink phase, where it formulated and rigorously tested assumptions about potential connections, ultimately rejecting unsupported hypotheses with clear reasoning.

\section{Related Work}

\subsection{Pre-training and Fine-tuning for Graphs}

The success of foundation models has inspired graph researchers to adopt a \emph{pre-train–then-fine-tune} paradigm.
Early efforts focused on self-supervised learning for graphs, where models such as GraphMAE~\citep{houGraphMAESelfsupervisedMasked2022a,houGraphMAE2DecodingenhancedMasked2023a}, GraphCL~\citep{yingTransformersReallyPerform2021}, DGI~\citep{velickovicDeepGraphInfomax2019}, GCC~\citep{qiuGCCGraphContrastive2020}, and GCA~\citep{zhuDeepGraphContrastive2020} are pre-trained on large-scale graph corpora and then fine-tuned for downstream tasks.
More recent approaches explore \emph{graph prompting}, where general-purpose pre-trained GNNs are adapted via textual or task-oriented prompts—for example, All-in-One~\citep{sunAllOneMultiTask2023a} and GraphPrompt~\citep{liuGraphPromptUnifyingPretraining2023a}. However, these methods remain constrained by the inherent architectural limitations of GNNs. As a result, their transferability is often limited to in-domain tasks and typically requires task-specific fine-tuning or additional parameters for optimal performance.

\subsection{LLMs for Graph Learning}
\paragraph{Graph-to-Text.}
Several studies transform subgraphs into natural-language prompts for LLMs~\citep{chenExploringPotentialLarge2024a,liuEvaluatingLargeLanguage2023a,wangCanLanguageModels2023a}.
However, subsequent analyses have found that ignoring structural information significantly degrades performance~\citep{huangCanLLMsEffectively2024}.

\paragraph{LLMs as Feature Enhancers.}
A common strategy is to leverage LLMs to embed heterogeneous node and edge attributes into a unified semantic space~\citep{yeLanguageAllGraph2024,yuLeveragingLargeLanguage2025,chenLabelfreeNodeClassification2024}.
For instance, OFA~\citep{liuOneAllTraining2024a} verbalises graph metadata and encodes it into dense language embeddings that augment the graph with enriched features.
ZeroG~\citep{liZeroGInvestigatingCrossdataset2024a} and OpenGraph~\citep{xiaOpenGraphOpenGraph2024} adopt similar approaches.
Nonetheless, these methods often depend on downstream predictors—typically graph neural networks (GNNs)—and are thus limited in the range of tasks they can effectively support.

\paragraph{LLM as Unified Predictor.}
An emerging line of research treats the LLM itself as the task head, bypassing traditional graph-specific predictors.
GraphGPT~\citep{tangGraphGPTGraphInstruction2024a} and GOFA~\citep{kongGOFAGenerativeOneForAll2024a} align graph embeddings with the LLM embedding space and apply instruction tuning for downstream adaptation.
UniGraph~\citep{heUniGraphLearningUnified2025} and TEA-GLM~\citep{wangLLMsZeroshotGraph2024b} introduce lightweight projection modules to enable zero-shot generalisation, while LLaGA~\citep{chenLLaGALargeLanguage2024b} tokenises entire graphs directly for LLM-based inference.
While these approaches are promising, embedding alignment can incur information loss, and relying solely on answer-only decoding often under-utilises the full reasoning capabilities of LLMs.

\subsection{Reasoning on Graphs}

Recent work has begun to assess the reasoning ability of LLMs on graph-structured problems.
GPT4Graph~\citep{guoGPT4GraphCanLarge2023b} evaluates GPT-4 on algorithmic tasks such as connectivity and max-flow, revealing encouraging results but limited scalability.
NLGraph~\citep{wangCanLanguageModels2023a} proposes a broad benchmark, showing that while LLMs manage simple instances, they struggle with structural complexity; instruction tuning offers only marginal improvements.
GraphWiz~\citep{chenGraphWizInstructionfollowingLanguage2024a} focuses on algorithmic reasoning (e.g., shortest paths), but omits standard learning tasks.
InstructGraph~\citep{wangInstructGraphBoostingLarge2024a} enhances supervised learning with natural-language instructions, yet falls short on cross-task generalisation. We introduce the first LLM-based framework to jointly integrate reinforcement learning and explicit reasoning, aiming to generalise across diverse graph tasks.

\section{Conclusion}
We presented \textsc{Graph-R1}, a GNN-free paradigm that formulates graph learning tasks—such as node classification, link prediction, and graph classification—as textual reasoning problems solvable by Large Reasoning Models (LRMs). To support this, we introduced the first reasoning dataset for graph machine learning, featuring detailed reasoning traces. Guided by task-specific \textit{rethink} templates, \textsc{Graph-R1} enables LRMs to reason over linearized graph structures. Extensive experiments show that \textsc{Graph-R1} outperforms strong baselines in zero-shot settings while producing interpretable predictions that expose its reasoning process. Our results highlight the promise of explicit reasoning for graph learning and open new directions at the intersection of graph learning and LRMs.

\section*{Limitations}
While \textsc{Graph-R1} shows strong zero-shot generalization and produces inherently interpretable reasoning across diverse graph tasks, it faces challenges when scaling to very large graphs. Current Large Reasoning Models (LRMs) have input length constraints, and linearizing large or complex graphs may exceed their context window. Future work may explore more efficient encoding methods to improve scalability.

\bibliography{graphreasoning}

\clearpage
\appendix

\section{Details of Datasets}
\label{appendix:dataset}

\begin{table*}[]
\centering
\begin{threeparttable}
\resizebox{0.9\textwidth}{!}{%
\begin{tabular}{cccccc}
\toprule
\textbf{Domain}                  & \textbf{Dataset} & \textbf{Avg.\#Nodes} & \textbf{AVG.\#Edges} & \textbf{\#Classes} & \textbf{\#Graphs} \\ \hline
 Social Network                   & Instagram        & 11339                & 155349               & 2                  &1   \\ \hline
 Web Link                         & WikiCS           & 11701                & 216123               & 10                 &1                 \\ \hline
 Logical Graph                    & Expla\_Graph     & 5.17                 & 4.25                 & -                  &2766              \\ \hline
\multirow{2}{*}{Knowledge Graph} & FB15K237         & 14541                & 310116               & 237                & 1                 \\
                                 & WN18RR           & 40943                & 93003                & 11                 & 1                 \\ \hline
\multirow{4}{*}{Citation}        & Arxiv            & 169343               & 1166243              & 40                 & 1                 \\
                                 & Citeseer         & 3186                 & 8554                 & 6                  & 1                 \\
                                 & Cora             & 2708                 & 10556                & 7                  & 1                 \\
                                 & Pubmed           & 19717                & 88648                & 3                  & 1                 \\ \hline
\multirow{5}{*}{E-commerce}      & Children         & 76875                & 1554578              & 24                 & 1                 \\
                                 & Computer         & 87229                & 721081               & 10                 & 1                 \\
                                 & Photo            & 48362                & 500939               & 12                 & 1                 \\
                                 & Products         & 54025                & 144638               & 47                 & 1                 \\
                                 & Sports           & 173055               & 1773500              & 13                 & 1                 \\ \hline

\multirow{7}{*}{Molecular}       & CHEMBL           & 25.87                & 55.92                & 1048               & 23874346          \\
                                 & BBBP             & 24.06                & 51.91                & 2                  & 2039              \\
                                 & ESOL             & 13.29                & 27.35                &                    -& 1128              \\
                                 & Freesolv         & 8.72                 & 16.76                &                    -& 642               \\
                                 & HIV              & 25.51                & 54.94                & 2                  & 41127             \\
                                 & Lipo             & 27.04                & 59                   &                    -& 4200              \\
                                 & PCBA             & 25.97                & 56.20                & 128                & 34017170          \\ \bottomrule
\end{tabular}%
}
\caption{Datasets Statistics (the "-" means that it is not appropriate to use the number of classes description. This is because Esol, Freesolv, Lipo is regression tasks, Expla\_graph is a Q-A task).}
\label{dataset statistic}
\end{threeparttable}
\end{table*}

All of the public datasets used in our paper were previously published, covering a multitude of domains. We report the detailed statistics for each dataset in Table~\ref{dataset statistic}. The detailed descriptions of these datasets are listed in the following:

\paragraph{Arxiv}
Arxiv~\citep{huOpenGraphBenchmark2020} is a large-scale citation graph derived from arXiv Computer Science papers. Each node corresponds to a paper and edges represent citation links between papers. The task is to classify each paper into one of 40 arXiv subcategories, such as "cs.LG" or "cs.AI". This dataset serves as a representative benchmark for large-scale node classification.

\paragraph{Citeseer}
The Citeseer~\citep{yangRevisitingSemisupervisedLearning2016} dataset is a citation network comprising research papers and their citation relationships within the computer science domain. Each node represents a research paper, and each edge signifies a citation relationship between two papers.

\paragraph{Cora}
The Cora~\citep{wenAugmentingLowresourceText2023} dataset is a citation graph where each node corresponds to a research paper, and each edge represents a citation link between papers. The dataset focuses on papers within the machine learning domain and includes 70 fine-grained categories, making the classification task particularly difficult.

\paragraph{Pubmed}
Pubmed~\citep{heHarnessingExplanationsLLMtoLM2024} is a citation network of biomedical research papers from the PubMed database. Each node is a paper and edges correspond to citation links. The classification task involves assigning each paper to one of three disease-related categories.

\paragraph{Children}
The Children~\citep{yanComprehensiveStudyTextattributed2023} dataset is a co-purchased or co-viewed product graph focused on children’s books. Nodes correspond to individual books, and edges connect books that were frequently browsed or bought together. Each node is associated with textual information including the book’s title and descriptive metadata.

\paragraph{Computer}
The Computer~\citep{yanComprehensiveStudyTextattributed2023} dataset is co-purchased or co-viewed product graph, where each node represents a product in the computer category, and edges indicate that two products were frequently co-purchased or co-viewed by users. The textual content associated with each node consists of user-generated reviews for the corresponding product.

\paragraph{Photo}
The Photo~\citep{yanComprehensiveStudyTextattributed2023} dataset is an e-commerce product graph where nodes represent photographic products, and edges indicate that two items were either co-purchased or co-viewed by users. The textual content of each node consists of user reviews associated with the corresponding product.

\paragraph{Products}
The Photo~\cite{heHarnessingExplanationsLLMtoLM2024} dataset is an e-commerce product graph where nodes represent Amazon products, and edges indicate that two items were either co-purchased or co-viewed by users. The textual content of each node consists of user reviews associated with the corresponding product.

\paragraph{Sports}
The Sports~\cite{yanComprehensiveStudyTextattributed2023} dataset is a co-purchased or co-viewed product graph in the sports domain. Nodes represent sports-related products, and edges indicate that two items were often purchased or viewed together. The associated text for each node consists of the product’s title.

\paragraph{FB15K237}
FB15K237~\cite{liuOneAllTraining2024a} is a large-scale knowledge graph where each node represents an entity (e.g., a person, location, or object) and each edge corresponds to a relational triple connecting two entities. Textual content for nodes is constructed from entity names and relation descriptions.

\paragraph{WN18RR}
WN18RR~\cite{liuOneAllTraining2024a} is another knowledge graph extracted from WordNet. It contains 40,943 nodes and 93,003 relations where each node is an English word and each edge represents the relation between two words. 

\paragraph{WikiCS}
WikiCS~\citep{mernyeiWikiCSWikipediabasedBenchmark2020} is a web link network constructed from English Wikipedia articles related to computer science. Nodes are individual articles, and directed edges represent hyperlinks between them. The node text is the full content of each article.

\paragraph{CHEMBL}
ChEMBL~\cite{gaultonChEMBLLargescaleBioactivity2012} is a molecular graph dataset where each graph corresponds to a chemical compound. Nodes represent atoms, and edges denote chemical bonds. The textual information for each molecule is given by its SMILES (Simplified Molecular Input Line Entry System) representation.

\paragraph{BBBP}
The BBBP~\cite{wuMoleculeNetBenchmarkMolecular2017} dataset comes from a study focused on modeling and predicting the permeability of the blood-brain barrier. The BBBP dataset contains binary labels indicating whether a compound can penetrate the blood-brain barrier (BBB) or not.

\paragraph{ESOL}
The ESOL~\cite{withnallMatchedMolecularPair2018} dataset contains water-solubility data for chemical compounds. Each molecule is modeled as a graph, with node and edge structures corresponding to atoms and bonds. SMILES strings serve as the textual representation.

\paragraph{Freesolv}
Freesolv~\cite{casasnovasTheoreticalPKaCalculations2014} consists of molecular graphs used for estimating hydration free energy. Each molecule is modeled by a graph of atoms and bonds. The SMILES representation is used as the text-based molecular description.

\paragraph{HIV}
The HIV~\cite{wuMoleculeNetBenchmarkMolecular2017} dataset consists of molecular graphs representing candidate compounds for HIV treatment. Nodes denote atoms and edges are chemical bonds. Each molecule is described by its SMILES string.

\paragraph{Lipo}
Lipo~\citep{wuMoleculeNetBenchmarkMolecular2017} is a molecular dataset focused on lipophilicity prediction. Each molecule is represented 524
as a graph with atoms as nodes and bonds as edges. The SMILES string encodes each molecule’s 525
structure in text form.

\paragraph{PCBA}
PCBA~\citep{wuMoleculeNetBenchmarkMolecular2017} is a large-scale molecular dataset for virtual screening. Each graph is a molecule, modeled by atoms and bonds, with SMILES strings representing the underlying chemical structure.

\paragraph{Expla\_Graph}
Expla\_Graphs~\citep{heGretrieverRetrievalaugmentedGeneration2024} is a graph question answering dataset on commonsense concepts. Each graph in Expla\_Graphs contains commonsense concepts connected by its relation. 

\paragraph{Instagram}
Instagram~\citep{liGLBenchComprehensiveBenchmark2024} is a social graph in which each node represents a user, and edges denote social connections such as following relationships. The textual content associated with each node is extracted from users’ self-introductions or profile descriptions.

\section{Prompt Template}
\label{appendix:prompt}

For each specific task type $\tau$, we design an appropriate prompt template to guide the model in understanding and solving the corresponding graph reasoning task. Our prompt templates are systematically constructed and consist of three main components: task-specific and dataset-related prompt prefix and question template, and a format-constrained instruction template. The instruction template is further categorized into two variants: \textit{normal} and \textit{rethink}.

The prompt prefix provides necessary background information, context, and relevant details about the graph structure and node attributes. The question template then formulates the concrete prediction or reasoning objective for the current instance. The instruction template standardizes the output format, ensuring consistency and clarity in model responses. 

A comprehensive overview of all prompt templates used for different task types is provided as follows: Table~\ref{tab:node_link_templates} shows prompt templates of node/link classification, Table~\ref{tab:graph_class_templates} shows prompt templates of graph classification, Table~\ref{tab:link_pred_templates} shows prompt templates of link prediction, and Table~\ref{tab:graph_reg_templates} shows prompt templates of graph regression.

In addition, we present a dedicated prompt template for summarizing node descriptions within graph reasoning tasks. This template is designed to effectively capture and condense the essential attributes and contextual information of individual nodes, facilitating more accurate and interpretable reasoning by the model. The detailed design of the node summary prompt template is provided in Table~\ref{tab:summary_prompt}.

\begin{table*}[htbp]
\centering
\captionsetup{width=0.9\textwidth}
\scriptsize
\begin{tabularx}{0.98\textwidth}{@{}p{3.6cm} X@{}}
\toprule
\textbf{Template Name} & \textbf{Content} \\
\midrule
Summary &
summary each node's content in no more than 25 words.\newline
Your response should strictly be in forms as follows: \newline
nodex:\textless{}your summary\textgreater{}\newline
eg:\newline
node1:optimality of myopic sensing in multi channel opportunistic access\newline
\{node descriptions\}\\
\bottomrule
\end{tabularx}
\caption{Prompt Templates in summarizing node descriptions}
\label{tab:summary_prompt}
\end{table*}

\begin{table*}[htbp]
\centering
\captionsetup{width=0.9\textwidth}
\scriptsize
\begin{tabularx}{0.98\textwidth}{@{}p{3.6cm} X@{}}
\toprule
\textbf{Template Name} & \textbf{Content} \\
\midrule
Prompt Prefix &
Classify the \textless{}target: essay / book / electronic product / user / product / fitness-related item / wikipedia page\textgreater{} represented by node \textless{}node\_id\textgreater{} using its subgraph data (text attributes and connections) as follows: \newline
Node description: \textless{}node description\textgreater{} \newline
Connection relationship among the nodes: \textless{}connection\textgreater{} \\
\midrule
Question Format &
Consider both semantic and structural information. Select strictly from: \{labels\}. Respond only with the category name and briefly summarize the reasoning process. \\
\midrule
Normal Instruction &
Your reasoning and response should be streamlined and restricted to within 2048 tokens. \newline
Your response should be in forms as follows: \newline
\quad Answer: your\_answer (e.g., \{sample\_answer\}) \newline
\quad Brief\_reasoning: your\_brief\_reasoning \\
\midrule
Rethink Instruction &
You must conduct reasoning inside \textless{}think\textgreater{}...\textless{}/think\textgreater{}. \newline
Inside \textless{}think\textgreater{}...\textless{}/think\textgreater{}, you should include: \newline
\quad - Structure information: \textless{}structure\textgreater{}...\textless{}/structure\textgreater{} \newline
\quad - Semantic similarities: \textless{}semantic\textgreater{}...\textless{}/semantic\textgreater{} \newline
After structure and semantic analysis, you msut provide \{candidate\} candidate answers with brief reasoning inside \textless{}comprehensive\textgreater{}...\textless{}/comprehensive\textgreater{}. \newline
Then, you must conduct re-reasoning inside \textless{}rethink\textgreater{}...\textless{}/rethink\textgreater{}. In this section, you should detailed consider each of your candidate answers as if they were the correct answer and evaluate their feasibility. \newline
After re-reasoning, you must conduct your final answer based on your above analysis. \newline
Finally, besides your reasoning, give your final response.\newline
Your full response must follow this format: \newline
\textless{}think\textgreater{} \newline
\quad \textless{}structure\textgreater{}Here show your structure analysis\textless{}/structure\textgreater{} \newline
\quad \textless{}semantic\textgreater{}Here show your semantic analysis\textless{}/semantic\textgreater{} \newline
\quad \textless{}comprehensive\textgreater{}Here show your comprehensive reasoning and list your candidate answers\textless{}/comprehensive\textgreater{} \newline
\quad \textless{}rethink\textgreater{}Here ongoing re-reasoning with each of your candidate answers inversely\textless{}/rethink\textgreater{} \newline
\quad Here show your final reasoning and answers \newline
\textless{}/think\textgreater{} \newline
Answer: your\_answer (e.g., \{sample\_answer\}) \newline
Brief\_reasoning: your\_brief\_reasoning \\
\bottomrule
\end{tabularx}
\caption{Prompt Templates of Node/Link Classification}
\label{tab:node_link_templates}
\end{table*}

\begin{table*}[htbp]
\centering
\captionsetup{width=0.9\textwidth}
\scriptsize
\begin{tabularx}{0.98\textwidth}{@{}p{3.6cm} X@{}}
\toprule
\textbf{Template Name} & \textbf{Content} \\
\midrule
Prompt Prefix &
Determine whether the chemical compound represented by the following molecular graph (nodes with atomic features and bond relationships) is predicted to exhibit activity (effectiveness) in each of the provided bioassays. \newline
Bioassays descriptions: \textless{}bioassays\_descriptions\textgreater{} \newline
Node description: \textless{}node description\textgreater{} \newline
Connection relationship among the nodes: \textless{}connection\textgreater{} \\
\midrule
Question Format &
Your response must include: A sequence of strict ‘Yes’ or ‘No’ answers for each property in order, separated by spaces (e.g., \{sample\_answer\}), and a concise explanation for your choices, referencing important structural features and the biological assay context. \\
\midrule
Normal Instruction &
Your reasoning and response should be streamlined and restricted to within 2048 tokens. \newline
Your response should be in forms as follows: \newline
\quad Answer: your\_answer (e.g., \{sample\_answer\}) \newline
\quad Brief\_reasoning: your\_brief\_reasoning \\
\midrule
Rethink Instruction &
- Same reasoning and output format as Node/Link Classification. \newline
- Only the task context differs; follow the steps and output structure above. \\
\bottomrule
\end{tabularx}
\caption{Prompt Templates of Graph Classification}
\label{tab:graph_class_templates}
\end{table*}

\begin{table*}[htbp]
\centering
\captionsetup{width=0.9\textwidth}
\scriptsize
\begin{tabularx}{0.98\textwidth}{@{}p{3.6cm} X@{}}
\toprule
\textbf{Template Name} & \textbf{Content} \\
\midrule
Prompt Prefix &
Classify the relationship between two \textless{}target: essays / books / electronic products / products / fitness-related items / wikipedia pages/entities\textgreater{} denoted as node \textless{}node\_id\textgreater{} and node \textless{}node\_id\textgreater{}, using the union of their corresponding subgraph (text attributes and connections) as follows: \newline
Node description: \textless{}node description\textgreater{} \newline
Connection relationship among the nodes: \textless{}connection\textgreater{} \\
\midrule
Question Format &
Consider: semantic and structural information. In your reasoning process provide the predicted connection bond value of the two target nodes between 0 and 1, set the threshold to 0.5. Based on your predicted connection bond value select strictly from: \hspace{1em}‘Yes, they have \{target: citation/co-purchased or co-viewed\} relationships’ \hspace{1em}or\hspace{1em}‘No, they do not have \{target: citation/co-purchased or co-viewed\} relationships’. Respond only with the choice content and briefly summarize the reasoning process. \\
\midrule
Normal Instruction &
Provide an estimated connection bond value (ranging from 0 to 1). A higher value indicates a stronger likelihood of a relationship. Consider multiple factors, such as: \newline
\quad - Structural information: Evaluate the direct and indirect connections between the two target nodes through their neighbors. \newline
\quad - Semantic similarities: Analyze the relevance or similarity in meaning between the two target nodes. \newline
\quad - Comprehensive information: If there exist two nodes that are semantically similar to each other, and these two nodes are respectively connected to the two target nodes, this can indirectly indicate the strength of the connection between the target nodes. \newline
Your reasoning and response should be streamlined and restricted to within 2048 tokens. \newline
Your response should follow this format: \newline
\quad Answer: your\_answer \newline
\quad Brief\_reasoning: your\_brief\_reasoning \newline
\quad Bond\_value: your\_predicted\_bond\_value \\
\midrule
Rethink Instruction &
Provide an estimated connection bond value (ranging from 0 to 1). A higher value indicates a stronger likelihood of a relationship. Consider multiple factors, such as: \newline
\quad - Structural information: Evaluate the direct and indirect connections between the two target nodes through their neighbors. \newline
\quad - Semantic similarities: Analyze the relevance or similarity in meaning between the two target nodes. \newline
\quad - Comprehensive information: If there exist two nodes that are semantically similar to each other, and these two nodes are respectively connected to the two target nodes, this can indirectly indicate the strength of the connection between the target nodes. \newline
If you can identify direct or indirect connections based on structural information, set the bond strength to 1 and specify the path(s) of connection in your reasoning. \newline
If no such connections can be identified, evaluate the bond strength based on the semantic similarity between the target node and its neighboring nodes semantics. \newline
You must conduct reasoning inside \textless{}think\textgreater{}...\textless{}/think\textgreater{}.\newline
Inside  \textless{}think\textgreater{}...\textless{}/think\textgreater{}, you should include: \newline
\quad - Structure information within  \textless{}structure\textgreater{}...\textless{}/structure\textgreater{} \newline
\quad - Semantic similarities within  \textless{}semantic\textgreater{}...\textless{}/semantic\textgreater{} \newline
\quad After structure and semantic analysis, provide comprehensive information inside \textless{}comprehensive\textgreater{}...\textless{}/comprehensive\textgreater{} \newline
\quad Then, you must conduct re-reasoning inside  \textless{}rethink\textgreater{}...\textless{}/rethink\textgreater{} . In this section, you should detailed consider each of the two given answers as if they were the correct answer and evaluate their feasibility. \newline
After re-reasoning, you must conduct your final answer based on your above analysis.\newline
Finally, besides your reasoning, give your final response.\newline
Your full response must follow this format: \newline
\textless{}think\textgreater{}\newline
\textless{}structure\textgreater{}Here show your structure analysis\textless{}/structure\textgreater{}\newline
\textless{}semantic\textgreater{}Here show your semantic analysis</semantic\textgreater{}\newline
\textless{}comprehensive\textgreater{}Here show your comprehensive reasoning\textless{}/comprehensive\textgreater{}\newline
\textless{}rethink\textgreater{}Here ongoing re-reasoning with each of the two candidate answers inversely\textless{}/rethink\textgreater{}\newline
Here show your final reasoning and answers\newline
\textless{}/think\textgreater{}\newline
Answer: your\_answer \newline
Brief\_reasoning: your\_brief\_reasoning \newline
Bond\_value: your\_predicted\_bond\_value \\
\bottomrule
\end{tabularx}
\caption{Prompt Templates of Link Prediction}
\label{tab:link_pred_templates}
\end{table*}

\begin{table*}[htbp]
\centering
\captionsetup{width=0.9\textwidth}
\scriptsize
\begin{tabularx}{0.98\textwidth}{@{}p{3.6cm} X@{}}
\toprule
\textbf{Template Name} & \textbf{Content} \\
\midrule
Prompt Prefix &
Calculate the chemical relevant properties using the given molecular graph (nodes with atomic features and bond relationships) as the following calculation requirements. \newline
Calculation requirements: \newline
\{description\} \newline
Calculate the \{target\} of this molecule. \newline
Node description: \textless{}node description\textgreater{} \newline
Connection relationship among the nodes: \textless{}connection\textgreater{} \\
\midrule
Question Format &
Your Response Must Include: A numerical answer, and the mathematical solution process, referencing important structural features and the biological assay context. \\
\midrule
Normal Instruction &
You are a chemistry expert assistant specialized in molecular graph regression tasks. \newline
Given a molecular graph with atomic features and bond relationships, you are asked to approximate the target value using the formula mentioned in calculation requirement
 \newline
Your task is to: \newline
\quad - Analyze the molecular structure based on the provided nodes and edges. \newline
\quad - Identify key chemical features that influence the target value (e.g., number and position of Cl atoms, ring systems, stereochemistry, hydrogen bonding capability). \newline
\quad - Estimate the target value based on the formula. \newline
\quad - Provide a final numeric prediction rounded to two decimal places. \newline
Please adjust the units of your final result so that the numerical value falls within the range of -30 to 30. \newline
Round the result to two decimal places. \newline
Respond strictly in the following format: \newline
\quad Answer: your\_answer (keep two decimal places, e.g., \{sample\_answer\}) \newline
\quad Brief\_reasoning: your\_brief\_reasoning \\
\midrule
Rethink Instruction &
You are a chemistry expert assistant specialized in molecular graph regression tasks. \newline
Given a molecular graph with atomic features and bond relationships, you are asked to approximate the target value using the formula mentioned in calculation requirements. \newline
Your task is to: \newline
\quad - Analyze the molecular structure based on the provided nodes and edges. \newline
\quad - Identify key chemical features that influence the target value (e.g., number and position of Cl atoms, ring systems, stereochemistry, hydrogen bonding capability). \newline
\quad - Estimate the target value based on the formula. \newline
\quad - Provide a final numeric prediction rounded to two decimal places. \newline
You must conduct reasoning inside \textless{}think\textgreater{}...\textless{}/think\textgreater{}. \newline Inside\textless{}think\textgreater{}...\textless{}/think\textgreater{}, you should include: \newline
\quad - Structure information within \textless{}structure\textgreater{}...\textless{}/structure\textgreater{} \newline
\quad - Semantic similarities within \textless{}semantic\textgreater{}...\textless{}/semantic\textgreater{} \newline
\quad After structure and semantic analysis, you must provide the range of target with brief reasoning inside \textless{}comprehensive\textgreater{}...\textless{}/comprehensive\textgreater{}  \newline
\quad Then, you must conduct re-reasoning inside \textless{}rethink\textgreater{}...\textless{}/rethink\textgreater{}. In this section, you should detailed consider your target range as if it were the correct range and evaluate its feasibility. \newline
After re-reasoning, you must conduct your final answer based on your above analysis. \newline
Finally, besides your reasoning, give your final response. \newline
Please adjust the units of your final result so that the numerical value falls within the range of -30 to 30. \newline
Round the result to two decimal places. \newline
Your full response must follow this format: \newline
\textless{}think\textgreater{}\newline
\textless{}structure\textgreater{}Here show your structure analysis\textless{}/structure\textgreater{}\newline
\textless{}semantic\textgreater{}Here show your semantic analysis</semantic\textgreater{}\newline
\textless{}comprehensive\textgreater{}Here show your comprehensive reasoning\textless{}/comprehensive\textgreater{}\newline
\textless{}rethink\textgreater{}Here ongoing re-reasoning with each of the two candidate answers inversely\textless{}/rethink\textgreater{}\newline
Here show your final reasoning and answers\newline
\textless{}/think\textgreater{}\newline
Answer: your\_answer (keep two decimal places, e.g., \{sample\_answer\}) \newline
Brief\_reasoning: your\_brief\_reasoning \\
\bottomrule
\end{tabularx}
\caption{Prompt Templates of Graph Regression}
\label{tab:graph_reg_templates}
\end{table*}

\section{Details of Implementation}
\label{appendix:experiment}

\paragraph{Datasets}
For the construction of our graph reasoning dataset, we initially collected 348{,}000 instances from 11 diverse graph datasets: \textit{Arxiv}, \textit{Citeseer}, \textit{Pubmed}, \textit{Instagram}, \textit{Children}, \textit{Computer}, \textit{Photo}, \textit{Sports}, \textit{Chemblpre}, \textit{Chempcba}, and \textit{Wn18rr}. Following the data filtering procedures described in Section~\ref{sec:datamethod}, we curated a high-quality subset comprising 10{,}000 instances, which serves as the training datasets for instruction fine-tuning and reinforcement learning. For evaluation, we adopt the datasets reported in the GOFA paper, and construct evaluation datasets by inserting into our prompt templates to ensure consistency and comparability.


\paragraph{Baselines Details}
For all baseline methods, we report the results as provided in the GOFA paper. Since our evaluation datasets are constructed to be consistent with those reported in GOFA, the results are directly comparable and ensure a fair evaluation.

\paragraph{Details of \textsc{Graph-R1}}
Graph-R1 is developed based on DeepSeek-R1-distilled-Qwen2.5-14B. We employ a two-stage training pipeline: supervised instruction fine-tuning followed by reinforcement learning with the rethink template. In supervised instruction fine-tuning, We utilize the LLaMA-Factory toolkit\citep{llamafactory} to perform full-parameter supervised fine-tuning. The learning rate is set to 1e-5, and all other hyperparameters follow their default settings. For the reinforcement learning stage, we adopt the GRPO training strategy implemented via the Tiny-Zero toolkit\citep{tinyzero}. For GRPO, wo use a batch size of 64, set learning rate to 1e-6, and apply the number of sampled outputs per question of 5. All other parameters are set to their default values. For inference, we utilize the VLLM~\citep{kwon2023efficient} framework to enable efficient batched decoding. All experiments are conducted on a cluster equipped with 8 NVIDIA A100 GPUs.

\section{some Cases from Graph-R1}
\label{appendix:case}

We provide some interesting cases from \textsc{Graph-R1}{'}s responses to illustrate the interpretability and principles behind its reasoning, as shown in Table~\ref{tab:case_studies}.

\begin{table*}[t]
  \centering
  \renewcommand{\arraystretch}{1.8} 
  \scriptsize 
  \begin{adjustbox}{max width=\textwidth} 
  \begin{tabular}{@{}p{2.5cm}p{3cm}p{3cm}p{3cm}p{3cm}@{}}
    \toprule
    \textbf{Task Type} & \textbf{Structure} & \textbf{Semantic} & \textbf{Comprehensive} & \textbf{Rethink} \\ \midrule
    Node Classification & 
    Central node (node0) connects to multiple nodes in a distributed system context. \newline
    \textbf{Details:} Node0 connects to node1, node3, node6, and node7. Node7 links to node2, node3, node4, and node0. Other connections include node5 to node6 and node4 to node7. &
    Focuses on distributed systems, load balancing, and consensus algorithms. \newline \textbf{Details:} Keywords like "distributed load balancing," "plurality consensus," "majority dynamics," and "multi-agent systems" dominate. Node4 mentions distributed learning dynamics, linking to machine learning. &
    Identifies three candidates based on structure and semantics. \newline \textbf{Details:} \newline
    1. cs.DC (Distributed Computing): Core focus on distributed algorithms, load balancing, and protocols. \newline
    2. cs.MA (Multiagent Systems): Multi-agent dynamics mentioned in node2 and node7. \newline
    3. cs.LG (Machine Learning): Distributed learning dynamics in node4. &
    Evaluates the feasibility of each candidate. \newline \textbf{Details:} \newline
    cs.DC: Most fitting due to focus on distributed systems and algorithms. \newline
    cs.MA: Relevant but secondary to distributed systems. \newline
    cs.LG: Specific to node4, less central than distributed computing\\ \midrule
    Link Prediction & 
    Evaluates direct and indirect connections between Node0 and Node1. \newline \textbf{Details}
    Node0 connects to nodes related to hashing methods (node2, node3) and genome processing (node7). Node1 connects to nodes focused on backscatter communication systems (node4, node8, node9). No direct or indirect links between the two target nodes. &
    Analyzes the thematic relevance between Node0 and Node1. \newline \textbf{Details:}
    Node0 focuses on OLAP for text analysis in humanities research, while Node1 is about MIMO SWIPT power optimization in wireless communication. No overlapping keywords or shared technical domains. &
    Considers intermediary nodes and shared themes to predict connection strength. \newline \textbf{Details:}
    Node0’s connections (node2, node3) focus on hashing methods and algorithms. Node1’s connections (node4, node8) focus on communication systems and throughput optimization. No shared intermediary nodes or overlapping application areas. &
    Tests hypotheses assuming citation relationships exist. \newline \textbf{Details:}
    If OLAP techniques were used in MIMO systems optimization, it would bridge the gap, but no evidence supports this. Alternatively, if MIMO SWIPT used OLAP for data analysis, the connection would still require shared terminology, which is absent. The domains remain distinct—humanities research vs. wireless communication engineering. \\
   
    \bottomrule
  \end{tabular}
  \end{adjustbox}
  \caption{Case studies for different graph tasks analyzed from structural, semantic, comprehensive, and rethink perspectives.}
  \label{tab:case_studies}
\end{table*}


\end{document}